\RequirePackage[svgnames]{xcolor}

\documentclass[11pt,a4paper]{mystyle}

\usepackage[all]{hypcap}
\usepackage[svgnames]{xcolor}
\usepackage[comma,authoryear,compress]{natbib}
\renewcommand{\cite}{\citep}

\usepackage{algorithm}
\usepackage{algorithmicx}
\usepackage{algpseudocode}
\usepackage{microtype}
\usepackage{booktabs}
\usepackage{float}
\usepackage{bigstrut}
\usepackage{multirow}

\usepackage{amsmath}
\usepackage{amssymb}
\usepackage{mathtools}
\usepackage{amsthm}
\usepackage{mathrsfs}
\usepackage{dsfont}
\usepackage{enumitem}
\usepackage{graphicx}
\usepackage{cleveref}
\usepackage{bxcoloremoji}

\usepackage{float}

\usepackage[utf8]{inputenc} %
\usepackage[T1]{fontenc}    %
\usepackage{hyperref}       %
\usepackage{url}            %
\usepackage{booktabs}       %
\usepackage{amsfonts}       %
\usepackage{nicefrac}       %
\usepackage{microtype}      %
\usepackage{amssymb}
\usepackage{fdsymbol}
\usepackage{wrapfig}
\usepackage{lipsum}
\usepackage{enumitem}
\usepackage{stackengine}
\usepackage[font=small,labelfont=bf]{caption}
\usepackage{color}
\usepackage{adjustbox}
\usepackage{soul}
\usepackage{rotating}
\usepackage{makecell}

\title{\textit{PhantomHunter:} Detecting Unseen Privately-Tuned LLM-Generated Text via Family-Aware Learning}

\runningtitle{\textit{PhantomHunter:} Detecting Unseen Privately-Tuned LLM-Generated Text via Family-Aware Learning}

\author[1,2]{\href{https://scholar.google.com/citations?user=xAw_fukAAAAJ}{\textcolor{black}{Yuhui Shi}}}
\author[1,2]{\href{http://undground.fun/}{\textcolor{black}{Yehan Yang}}}
\author[1]{\href{https://sheng-qiang.github.io/}{\textcolor{black}{Qiang Sheng}}}
\author[1,2]{\href{https://scholar.google.com/citations?user=RxULDkAAAAAJ}{\textcolor{black}{Hao Mi}}}
\author[1,2]{\href{https://beanandrew.github.io/}{\textcolor{black}{Beizhe Hu}}}
\author[1,2]{\href{https://xuchaoxi.github.io/}{\textcolor{black}{Chaoxi Xu}}}
\author[1,2]{\href{https://scholar.google.com/citations?user=fSBdNg0AAAAJ}{\textcolor{black}{Juan Cao}}}

\affil[1]{Media Synthesis and Forensics Lab, Institute of Computing Technology, Chinese Academy of Sciences}
\affil[2]{University of Chinese Academy of Sciences}

\emails{{\{\href{mailto:shiyuhui22s@ict.ac.cn}{shiyuhui22s},\href{mailto:shengqiang18z@ict.ac.cn}{shengqiang18z},\href{mailto:mihao24s@ict.ac.cn}{mihao24s},\href{mailto:hubeizhe21s@ict.ac.cn}{hubeizhe21s},\href{mailto:xuchaoxi@ict.ac.cn}{xuchaoxi},\href{mailto:caojuan@ict.ac.cn}{caojuan}\}@ict.ac.cn,\href{mailto:yyh4233@outlook.com}{yyh4233@outlook.com}}}

\begin{document}

\begin{abstract}
With the popularity of large language models (LLMs), undesirable societal problems like misinformation production and academic misconduct have been more severe, making LLM-generated text detection now of unprecedented importance.
Although existing methods have made remarkable progress, a new challenge posed by text from privately tuned LLMs remains underexplored.
Users could easily possess private LLMs by fine-tuning an open-source one with private corpora, resulting in a significant performance drop of existing detectors in practice.
To address this issue, we propose PhantomHunter, an LLM-generated text detector specialized for detecting text from unseen, privately-tuned LLMs.
Its family-aware learning framework captures family-level traits shared across the base models and their derivatives, instead of memorizing individual characteristics. 
Experiments on data from LLaMA, Gemma, and Mistral families show its superiority over 7 baselines and 3 industrial services, with F1 scores of over 96\%.
\vspace{7mm}

\coloremojicode{1F4C5} \textbf{Date}: June 18, 2025

\end{abstract}

\maketitle
\vspace{3mm}

\section{Introduction}

Large language models (LLMs) have successfully revolutionized the way people organize and produce text and inspire productivity in a wide range of applications~\cite{kaddour2023applications}. However, undesirable societal problems caused by the misuses of LLMs also rapidly emerge, such as cheating in academic writing~\cite{koike2024outfox}, creating misinformation~\cite{chen2024combating}, and accelerating information pollution~\cite{puccetti2024ai}. As the front-line barrier against such threats, automatic detection techniques of LLM-generated text (LLMGT) is now of unprecedented importance.

LLMGT detection generally distinguishes LLM-generated and human-written text via binary classification.
Existing methods either learn common textual features (e.g., stylistic clues; \citealp{Guo2023}) shared across LLMs using representation learning or design distinguishable metrics between human and LLM texts based on LLMs' internal signals (e.g., token probabilities;~\citealp{poger}).
For both categories, their tests were mostly conducted on data from publicly available LLMs, assuming that users generate text using public, off-the-shelf services.
\textbf{We argue that this situation is being changed due to the recent development of the open-source LLM community.}
With the help of platforms like HuggingFace\footnote{\url{https://huggingface.co/models}} and the efficient LLM training techniques like low-rank adaptation (LoRA)~\cite{lora}, building fine-tuned LLMs with customized private datasets has become much easier than before.
For instance, there have been over 60k Llama-based derivative models on HuggingFace~\cite{meta_ai_llama_2024}.
After private fine-tuning on unknown corpus, the learned characteristics of base models could change and the LLMGT detectors would fail (will empirically show in Section~\ref{sec:pre-experiment}), shaping a new risk that malicious users can generate harmful texts privately without being caught by LLMGT detectors. A new challenge arises: \textbf{How to detect text generated by privately-tuned open-source LLMs?}

To address this issue, we first conduct an analysis of differences between the base LLMs and its derivative models via fine-tuning on data of different topics.
The result reveals that compared to other non-homologous models, the token probability lists of text on the fine-tuned LLM have a higher similarity to the probability lists on the base LLM, exhibiting clear ``family trait''.
Inspired by this finding, we propose to perform family-aware learning by extracting probabilistic features of base models and design \textbf{PhantomHunter}, a detector targeted at text generated by unseen privately-tuned LLMs.
Our main contributions are as follows:
\begin{itemize}
    \item \textbf{New Risk:} We empirically expose the new risk that current LLM-generated text detectors may fail to detect text from privately-tuned LLMs. 
    \item \textbf{New Method:} We propose PhantomHunter, which trained via family-aware learning to learn common traits shared in typical open-source LLM families.
    \item \textbf{High Performance:} Experiments on data from 3 LLM families show that PhantomHunter outperforms 7 baselines and 3 industrial services.
\end{itemize}

\section{How Does Fine-tuning Affect LLMGT Detection?}
\label{sec:pre-experiment}
\begin{figure}[ht]
\begin{center}
\centerline{\includegraphics[width=0.6\columnwidth]{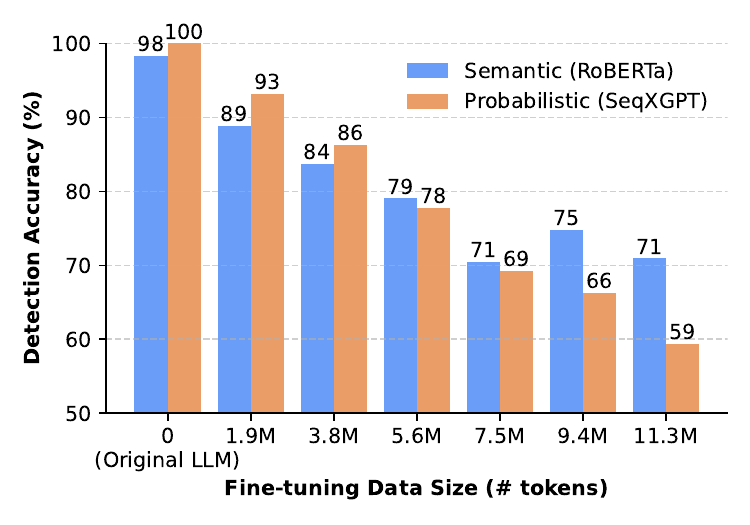}}
\caption{\label{fig:finetune-detection} Detection accuracy for LLM-generated text with increasing amounts of fine-tuning data.}
\end{center}
\end{figure}

\noindent{\textbf{Preliminary Experiment.}} To explore the extent to which fine-tuning affects the performance of LLM-generated text detectors, we fine-tuned LLaMA-2 7B using arXiv abstract corpus by saving checkpoints at different fine-tuning steps, which were then used to generated test samples. 
We experimented on two trainable detectors, RoBERTa (semantic-based; ~\citealp{Guo2023} and SeqXGPT (probability-based;~\citealp{wang2023seqxgpt}). Figure~\ref{fig:finetune-detection} displays \textbf{their decaying trends in accuracy as the fine-tuning corpus size increases}. For the LLM fine-tuned on 11.3M tokens, SeqXGPT even encounters an astonishing accuracy drop of 41\%.

\begin{figure}[ht]
\begin{center}
\centerline{\includegraphics[width=0.6\columnwidth]{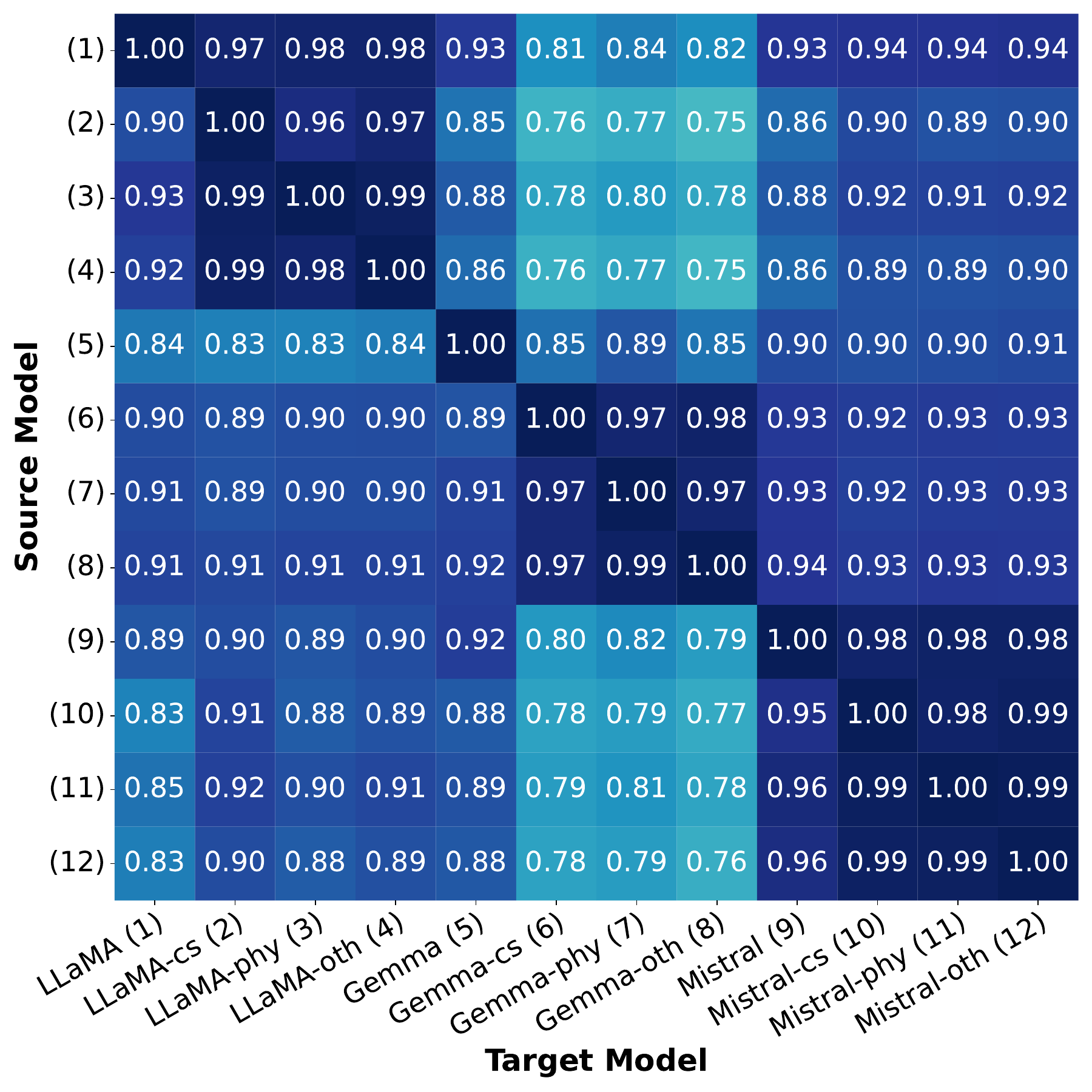}}
\caption{\label{fig:token-heatmap} Cosine similarity between the token probability lists of fine-tuned models and those of the base ones. Darker colors indicate higher similarity.}
\end{center}
\end{figure}

\noindent{\textbf{Impact of Fine-tuning on Features.}}
We fine-tuned 3 open-source LLMs using 3 sets of corpus from different domains, and obtained the generated text of 12 LLMs including the original base LLM (see Section~\ref{sec:dataset} for details). We obtain their probability lists on each LLM, and calculate cosine similarity between the probability lists of each fine-tuned model and that of the base one. From Figure~\ref{fig:token-heatmap}, we intuitively observe that the probability lists of LLMs from the same LLM are more similar to each other, and clearly different from those of the others, which shows significant ``family trait''.

\noindent{\textbf{Inspiration.}} 
This observation reveals that the probability lists of the base models could be useful for modeling unseen fine-tuned ones.
However, such ``family traits'' are not fully exploited by existing detectors because they treat all LLMs equally and may only learn the commonality across all involved LLMs (regarding the 12$\times$12 area in the heatmap), not the family-level commonality (regarding each 4$\times$4 areas), shaping a looser boundary against human text. Targetedly, we design PhantomHunter to learn such traits to address this issue.

\section{Proposed Method: PhantomHunter}
\begin{figure}[ht]
\begin{center}
\centerline{\includegraphics[width=\textwidth,trim=30 220 24 60,clip]{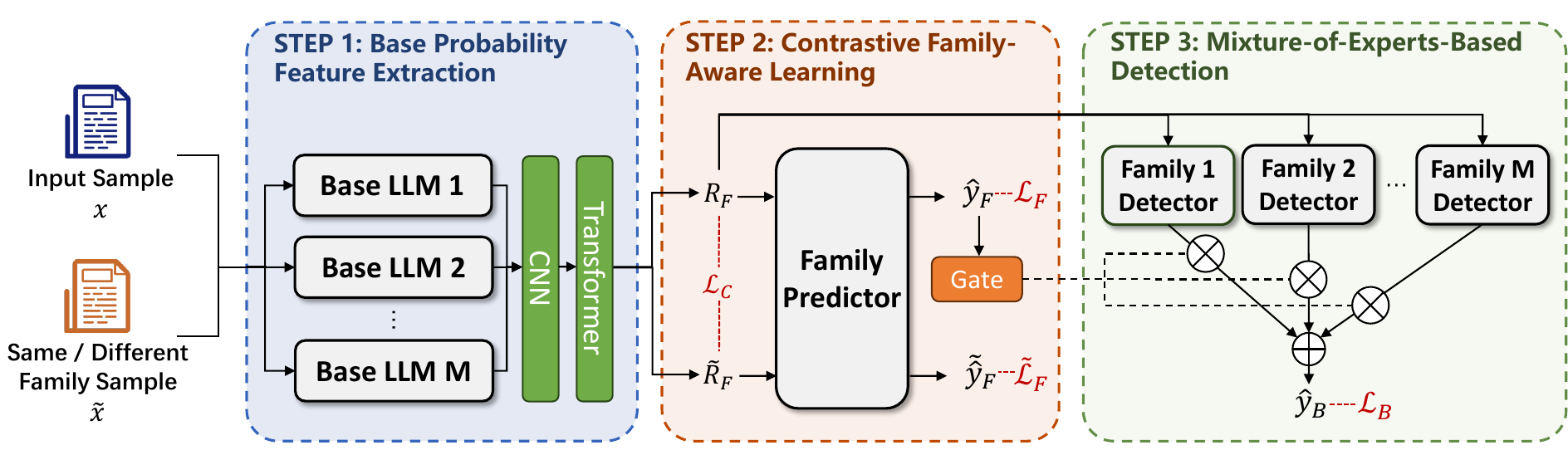}}
\caption{Overall architecture of PhantomHunter. Given a text sample $\mathbf{x}$, it \textbf{1)} extracts the probability feature from $M$ base models and encode them with CNN and transformer blocks; \textbf{2)} predicts the family of $\mathbf{x}$ to determine the family gating weights; and \textbf{3)} feeds the representation $\mathbf{R}_{F}$ to a mixture-of-experts network controlled by the gating weights from Step 2 for final prediction of $\mathbf{x}$ being LLM-generated. During training, a contrastive learning is applied in each mini-batch to better model family relationships.}
\label{fig:phantom-hunter}
\end{center}
\end{figure}

PhantomHunter is a family-aware learning-based detector for unseen privately-tuned LLM-generated texts. The key insight behind PhantomHunter is to model the inherent family-level features shared between base LLMs and their fine-tuned descendants. 
Building on these observations, PhantomHunter employs a family-aware learning approach to detect texts from unseen privately-tuned LLMs. As illustrated in Figure~\ref{fig:phantom-hunter}, the architecture consists of three main components: 1) a base probability feature extractor, 2) a contrastive learning-based family encoder, and 3) a mixture-of-experts detection module. We detail each component in the following subsections.

\subsection{Base Probability Feature Extraction}

For a given text sequence $\mathbf{x}$, we pass it through $M$ base LLMs $\theta_1, \theta_2, \ldots, \theta_M$ to obtain the probability lists $\mathbf{p} \in \mathbb{R}^{M \times N}$:
\begin{equation}
\mathbf{p} = (\mathbf{p}^{\theta_1}, \mathbf{p}^{\theta_2}, \ldots, \mathbf{p}^{\theta_M}),
\end{equation}
where $\mathbf{p}^{\theta_i} = (p^{\theta_i}_1, p^{\theta_i}_2, \ldots, p^{\theta_i}_N)$, $N$ is the text sequence length, $M$ is the number of base models, and $p^{\theta_i}_j$ represents the probability of generating token $x_j$ given context $x_{<j}$ using LLM $\theta_i$.
We then use convolutional neural networks and Transformer encoders to extract base probability features $\mathbf{R}_F \in \mathbb{R}^{M \times d}$, where $d$ is the feature dimension.

\subsection{Contrastive Family-Aware Learning}

To better model family relationships in the feature space and enhance the detector's generalization within the same family, we treat samples from the same family as augmentations and employ contrastive learning to enhance the representation $\mathbf{R}_F$.
Specifically, within each mini-batch, texts generated by models from the same family as the original text's model are treated as positive examples, while others are negative ones. We compute a SimCLR-style~\cite{Chen2020} contrastive loss $\mathcal{L}_C$:
\begin{equation}
\mathcal{L}_C = - \sum_{i \in b} \log  \frac{\exp(\delta(\mathbf{R}^m_{F_i}, \tilde{\mathbf{R}}^m_{F_i})/t)}{\sum_{j=1}^{2|b|}  \mathbf{1}_{[j \neq i]}  \exp(\delta(\mathbf{R}^m_{F_i}, \mathbf{R}_{F_j})/t)},
\end{equation}
where $\mathbf{R}^m_{F_i}$/$\tilde{\mathbf{R}}^m_{F_i}$ are the embeddings of the original/augmented sample, respectively, $\delta$ is a dot-product function, and $t$ controls the temperature.
We use a multi-layer perceptron (MLP) to classify the LLM family:
\begin{equation}
\hat{y}_F = \textrm{softmax}(\mathrm{MLP_F}(\mathbf{R}_F)),
\end{equation}
where $\hat{y}_F \in \{\theta_1, \theta_2, \ldots, \theta_M\}$ represents the predicted family.
\subsection{Mixture-of-Experts-Based Detection}

Finally, we employ a mixture-of-experts network for binary classification of human versus AI-generated text. Each expert detector specializes in detecting texts from specific LLM families, enhancing detection performance. Specifically, we use the family prediction result $\hat{y}_F$ as a gating signal to control the weighting of each expert's prediction, yielding the final binary judgement:
\begin{equation}
\hat{y}_B = \sum_{i=1}^{M} \hat{y}^i_F \cdot \textrm{softmax}(\mathrm{MLP_B}(\mathbf{R}_F)),
\end{equation}
where $\hat{y}^i_F$ represents the probability that the text belongs to the family of base model $\theta_i$ according to the family classifier. $\hat{y}_B \in [0,1]$ denotes the probability of the $\mathbf{x}$ being LLM-generated.

We optimize the PhantomHunter model with:
\begin{equation}
\mathcal{L} = \lambda_1(\mathcal{L}_F + \tilde{\mathcal{L}}_F) + \lambda_2\mathcal{L}_B + \lambda_3\mathcal{L}_C,
\end{equation}
where $\mathcal{L}_F$ and $\tilde{\mathcal{L}}_F$ represent the cross-entropy losses for family classification of $\mathbf{x}$ and $\tilde{\mathbf{x}}$ respectively, $\mathcal{L}_B$ is the cross-entropy loss for LLM text detection, $\mathcal{L}_C$ is the contrastive loss. $\lambda_i\ (i=1,2,3)$  are hyperparameters to balance the losses.

PhantomHunter captures inherent signatures that persist through fine-tuning by learning family-level features from observable fine-tuned models derived from the same base LLMs, enabling detection of unseen models from known families. Rather than focusing on individual model characteristics, we emphasize shared patterns defining model families. This family-aware approach allows the detector to generalize to unseen fine-tuned models by recognizing persistent probability patterns that remain after extensive fine-tuning. The method is effective because fine-tuning primarily adapts a model's behavior to specific domains while preserving much of the underlying probabilistic structure inherited from the base model.

\section{Experiment}
\label{sec:experiment}

We experimentally answer the following questions:

\noindent \textbf{EQ1}: How effective is PhantomHunter at detecting texts from unseen fine-tuned LLMs?

\noindent \textbf{EQ2}:  How do different components of PhantomHunter contribute to its overall performance?

\noindent \textbf{EQ3}:  How does PhantomHunter perform in  diverse real-world settings?

\subsection{Dataset Construction}
\label{sec:dataset}

\begin{table}[ht]
\small
\caption{\label{tab:finetune-data} Statistics of the corpora for fine-tuning.}
\centering
\begin{tabular}{lr}
\toprule
\textbf{ArXiv/Q\&A Domain} & \textbf{\# Tokens} \\
\midrule
Computer Science (cs) & 6,441,516 \\
Physics (phy) & 5,955,389 \\
Others (oth; q-bio/stat/eess/math/q-fin/econ) & 3,939,626 \\
\midrule
ELI5 & 202,658 \\
Finance (fin) & 234,787 \\
Medicine (med) & 254,292 \\
\bottomrule
\end{tabular}
\end{table}

We fully simulate the two most common usage scenarios of LLM: writing and question-answering.
For writing, we collect 69,297 abstracts of academic papers from arXiv archive\footnote{\url{https://arxiv.org/archive/}}, categorizing them by primary subjects (domains). And for Q\&A, we collect 3,062 Q\&A pairs in ELI5, finance, and medicine domains from HC3 dateset~\cite{Guo2023}.
Table~\ref{tab:finetune-data} shows the corpora size. Note that we merge subjects other than physics and computer science for arXiv part to balance fine-tuning scales across domains. 
We select open-source LLaMA-2 7B-Chat~\cite{llama2}, Gemma 7B-it~\cite{gemma}, and Mistral 7B-Instruct-v0.1~\cite{mistral} as base models.
For each scenario, we fine-tuned each base model using full-parameter fine-tuning (Full) and LoRA fine-tuning on the corresponding 3 domain-specific corpora, resulting in 9 derivative models. We then generate abstracts with 3 base models and 9 arXiv-based derivative models, and answers with the same base models and 9 Q\&A-based derivative ones. For the arXiv dataset, we designate LLMs fine-tuned on computer science (cs) data as unseen ones, reserving them exclusively for testing. And for the Q\&A dataset, we designate LLMs fine-tuned on finance (fin) data as unseen ones. Detailed statistics of our experimental dataset are shown in Table~\ref{tab:data-stat}.

\begin{table}[ht]
\small
\caption{\label{tab:data-stat} Statistics of the evaluation dataset for privately-tuned LLMGT detection. FT Domain: Domains of data for LLM fine-tuning (``base'' denotes no fine-tuning).}
\begin{center}
\begin{tabular}{ccc ccc r}
\toprule
& \textbf{Dataset} & \textbf{Source} & \multicolumn{1}{c}{\textbf{FT Domain}} & \textbf{\#Samples} \\
\midrule
\multirow{8}{*}{\rotatebox[origin=c]{90}{arXiv}}& \multirow{4}{*}{Train}
& LLaMA &base/phy/oth & 1,564/1,969/1,971\\
& & Gemma &base/phy/oth & 1,604/1,455/1,457  \\
& & Mistral &base/phy/oth & 1,778/1,877/1,826  \\
& & Human & \multicolumn{1}{c}{-} & 1,145 \\
\cmidrule{2-5}
& \multirow{4}{*}{Test}
& LLaMA & cs & 2,076 \\
& & Gemma &  cs &  1,540 \\
& & Mistral & cs & 2,002 \\
& & Human & \multicolumn{1}{c}{-} & 208 \\
\midrule
\multirow{8}{*}{\rotatebox[origin=c]{90}{Q\&A}} & \multirow{4}{*}{Train}
& LLaMA  & base/med/ELI5& 680/256/220\\
& & Gemma & base/med/ELI5 & 680/256/220 \\
& & Mistral & base/med/ELI5 & 680/256/220 \\
& & Human & \multicolumn{1}{c}{-} & 5,654 \\
\cmidrule{2-5}
& \multirow{4}{*}{Test}
& LLaMA & fin &  204\\
& & Gemma & fin &  204\\
& & Mistral & fin & 204 \\
& & Human & \multicolumn{1}{c}{-} & 612 \\
\midrule
\multicolumn{4}{c}{\textbf{Total}} & 32,818 \\
\bottomrule
\end{tabular}
\end{center}
\end{table}

\subsection{Experimental Setup}
\noindent\ul{\textbf{Implementation Details.}}
In PhantomHunter, the CNN encoder comprises three convolutional layers followed by max-pooling, and the Transformer encoder has two attention layers with four attention heads each.
The feature dimension $d$ is set to 128. For contrastive learning, we use a temperature $t$ of 0.07. The hyperparameters in the loss function are set as $\lambda_1 = 1.0$, $\lambda_2 = 1.0$, and $\lambda_3 = 0.5$, respectively.
The model is trained using the Adam optimizer with a learning rate of 2e-5 and batch size of 32. We train for 10 epochs and select the best checkpoint based on validation performance.
All experiments are conducted on a server with 4 NVIDIA A100 GPUs.

\noindent\ul{\textbf{Evaluation Metrics.}} We report F1 scores on each testing subsets with 0.5 as the threshold . For comparison with commercial detectors, we add the true positive rate under 1\% false positive rate to better align with practical scenarios~\cite{tufts2025}.

\noindent\ul{\textbf{Compared Baselines.}}
We select the following 7 methods for comparison:
\begin{itemize}
    \item \textbf{RoBERTa}~\cite{Guo2023}: A fine-tuned RoBERTa model~\cite{roberta} that uses semantic features to distinguish between human and AI-generated text.
    \item \textbf{T5-Sentinel}~\cite{chen2023token}: A model based on T5~\cite{t5} that treats token prediction as implicit classification to detect generated text.
    \item \textbf{SeqXGPT}~\cite{wang2023seqxgpt}: A white-box method that treats token probability lists as waveforms and applies CNN and attention networks for detection.In the experiments,We use mistral-7b~\cite{mistral}, Llama-2-7b-chat~\cite{llama2}, and gemma-7b~\cite{gemma} as white-box models to compute the probability lists.
    \item \textbf{DNA-GPT}~\citep{yang2024dnagpt}: A source-attribution method that leverages multiple regenerations to identify model-specific statistical fingerprints, functioning in both black-box and white-box settings. In the experiment, we adopted the white-box setting and selected Llama-2-7b-chat~\cite{llama2} as the proxy base model, with re-generations num N = 10, the truncation ratio $\gamma$ = 0.5, and max new tokens num t = 250. Since this method is training-free, we conducted a grid search for the classification threshold with a step size of 1e-4, and reported the best result as the upper performance bound of this method. 
    \item \textbf{DetectGPT}~\cite{mitchell2023detectgpt}: A zero-shot detection approach that identifies AI-generated text by comparing the probability of original passages against multiple perturbed variants without requiring model fine-tuning. In the experiment, we selected T5-3B~\cite{t5} as the mask-filling model, and Llama-2-7b-chat~\cite{llama2} as the proxy base model, with the number of perturbations as 100. In experiment report, we follow a similar implementation to DNA-GPT.
    \item \textbf{Fast-DetectGPT}~\cite{bao2023fast} : An efficient improvement over DetectGPT that introduces the concept of conditional probability curvature to identify machine-generated text, while significantly reducing computational overhead through optimized perturbation sampling strategies that maintain or enhance detection performance. In the experiment, we selected gpt-neo-2.7b~\cite{gpt-neo} as the sampling model and the scoring model.
    \item \textbf{DeTeCtive}~\cite{guo2024detective}: A comprehensive framework that reformulates the task from a binary classification problem to one of distinguishing various writing styles, treating each large language model as a unique author with specific stylistic patterns. DeTeCtive combines multi-level contrastive learning and multi-task auxiliary learning methods to enable fine-grained feature extraction that captures relationships between different text sources.In the experiments, We use SimCSE-RoBERTa-base~\cite{gao2021simcse} as the pretrain model. To preserve the original semantic space and prevent overfitting, we freeze the model's embedding layer during training.  During testing, we adopt a KNN-based retrieval strategy using the validation set to automatically determine the optimal number of neighbors K within the range of 1 to 5, which is then used for predicting out-of-distribution data.
\end{itemize}

\subsection{Main Results (EQ1)}
\label{sec:main-results}

Table~\ref{tab:main-results} presents the binary classification results (human vs. LLM-generated) of PhantomHunter and baselines on unseen fine-tuned models.
Compared to black-box baselines (RoBERTa, T5-Sentinel and DeTeCtive) and proxy-based white-box baselines (SeqXGPT, DNA-GPT, DetectGPT and Fast-DetectGPT), PhantomHunter demonstrates superior performance in detecting texts from unseen fine-tuned models, with F1 scores for both human and LLM text exceeding 90\% in each setting.
With full fine-tuning, PhantomHunter improves MacF1  score over the best baseline by 3.65\% and 2.96\% on both datasets, respectively; and with LoRA fine-tuning, the improvements are 2.01\% and 6.09\% respectively.
The result demonstrates PhantomHunter’s powerful detection capability for texts generated by unseen fine-tuned LLMs.

\begin{table}[ht]
    \centering
    \small
    \caption{\label{tab:main-results} F1 scores for unseen fine-tuned LLM-generated text detection. The two best results are \textcolor{black}{\textbf{bolded}} and \underline{underlined}. BFE: Base probability feature extraction. CL: Constrative learning. MoE: Mixture-of-experts.}
    \setlength{\tabcolsep}{3pt}
    \setlength{\aboverulesep}{0pt}
    \setlength{\belowrulesep}{0pt}
    \newcommand\astrut{\rule[8pt]{0pt}{2pt}}
    \newcommand\bstrut{\rule[-5pt]{0pt}{5pt}}
    \begin{tabular}{lcccccccccccc}
        \toprule
        \astrut\multirow{3}{*}[-0.3em]{\textbf{Method}} & \multicolumn{6}{c}{\textbf{arXiv}} & \multicolumn{6}{c}{\textbf{Q\&A}} \\
        & \multicolumn{3}{c}{\textbf{Full}} & \multicolumn{3}{c}{\textbf{LoRA}} & \multicolumn{3}{c}{\textbf{Full}} & \multicolumn{3}{c}{\textbf{LoRA}}\bstrut \\
        \cmidrule{2-13}
        \astrut& Human & Gen. & \cellcolor{yellow!30}Macro & Human & Gen. & \cellcolor{yellow!30}Macro &Human & Gen. & \cellcolor{yellow!30}Macro & Human & Gen. & \cellcolor{yellow!30}Macro \bstrut \\
        \midrule
        \astrut RoBERTa & 67.18 & 98.47 & \cellcolor{yellow!30}82.83 & 61.68 & 97.67 & \cellcolor{yellow!30}79.67 & 90.26 & 87.91 & \cellcolor{yellow!30}89.09 & 86.08 & 80.70 & \cellcolor{yellow!30}83.39 \\
        T5-Sentinel & 68.11 & 98.95 & \cellcolor{yellow!30}83.53 & 71.60 & 99.19 & \cellcolor{yellow!30}85.40 & 88.47 & 89.27 & \cellcolor{yellow!30}88.87 & 90.16 & 89.56 & \cellcolor{yellow!30}89.86 \\
        DeTeCtive & 69.66 & 99.04 & \cellcolor{yellow!30}84.35 & 85.83 & 99.39 & \cellcolor{yellow!30}92.61 & 93.36 & \underline{95.03} & \cellcolor{yellow!30}\underline{94.19} & 89.87 & 91.87 & \cellcolor{yellow!30}90.87 \\
        SeqXGPT & 86.24 & 99.46 & \cellcolor{yellow!30}92.85 & 87.18 & 99.46 & \cellcolor{yellow!30}93.32 & 92.20 & 94.74 & \cellcolor{yellow!30}93.47 & 87.95 & 85.71 & \cellcolor{yellow!30}86.83 \\
        DNA-GPT & 11.62 & 73.98 & \cellcolor{yellow!30}42.80 & 73.08 & 99.00 & \cellcolor{yellow!30}86.04 & 73.65 & 73.74 & \cellcolor{yellow!30}73.70 & 60.34 & 69.00 & \cellcolor{yellow!30}64.67 \\
        DetectGPT & 26.59 & 95.58 & \cellcolor{yellow!30}61.09 & 54.11 & 98.31 & \cellcolor{yellow!30}76.21 & 80.53 & 76.23 & \cellcolor{yellow!30}78.38 & 74.96 & 71.55 & \cellcolor{yellow!30}73.26 \\
        Fast-DetectGPT & 38.70 & 94.74 & \cellcolor{yellow!30}66.72 & \underline{90.82} & \underline{99.68} & \cellcolor{yellow!30}\underline{95.25} & 92.43 & 91.86 & \cellcolor{yellow!30}92.15 & 89.47 & 87.87 & \cellcolor{yellow!30}88.67 \bstrut \\
        \midrule
        \astrut \textbf{PhantomHunter} & \textcolor{black}{\textbf{92.75}} & \textcolor{black}{\textbf{99.73}} & \cellcolor{yellow!30}\textcolor{black}{\textbf{96.24}} & \textcolor{black}{\textbf{94.47}} & \textcolor{black}{\textbf{99.80}} & \cellcolor{yellow!30}\textcolor{black}{\textbf{97.14}} & \textcolor{black}{\textbf{97.30}} & \textcolor{black}{\textbf{96.58}} & \cellcolor{yellow!30}\textcolor{black}{\textbf{96.98}} & \textcolor{black}{\textbf{96.53}} & \textcolor{black}{\textbf{96.27}} & \cellcolor{yellow!30}\textcolor{black}{\textbf{96.40}} \\
        \ \ \textit{w/o} BFE & 63.81 & 98.33 & \cellcolor{yellow!30}81.07 & 66.24 & 98.10 & \cellcolor{yellow!30}82.17 & 90.94 & 88.93 & \cellcolor{yellow!30}89.93 & 87.43 & 83.21 & \cellcolor{yellow!30}85.32 \\
        \ \ \textit{w/o} CL & \underline{86.77} & \underline{99.56} & \cellcolor{yellow!30}\underline{93.16} & 88.89 & 99.54 & \cellcolor{yellow!30}94.21 & \underline{94.88} & 93.19 & \cellcolor{yellow!30}94.03 & 94.08 & 93.47 & \cellcolor{yellow!30}93.78 \\
        \ \ \textit{w/o} MoE & 81.20 & 99.21 & \cellcolor{yellow!30}90.20 & 88.60 & 99.54 & \cellcolor{yellow!30}94.07 & 93.96 & 92.24 & \cellcolor{yellow!30}93.10 & \underline{94.47} & \underline{94.08} & \cellcolor{yellow!30}\underline{94.27} \bstrut \\
        \bottomrule
    \end{tabular}
\end{table}

\subsection{Ablation Studies (EQ2)}
To evaluate the contribution of each component, we perform ablation studies by replacing base feature extractor to a RoBERTa (\textit{w/o} BFE), removing contrastive loss (\textit{w/o} CL), and removing mixture-of-experts components (\textit{w/o} MoE) from PhantomHunter respectively, and the results are shown in Table~\ref{tab:main-results}.

Removing any component leads to a degradation of detection performance, especially for BFE, since it is the probabilistic characterization of the base model that embodies the family commonality.
Without contrastive loss, the probability representations of different family-generated texts become less distinguishable in the feature space.
Removing the mixture-of-experts module forces the classifier to find a decision boundary between all family-generated texts and human texts, without specializing in particular family models.

\subsection{Further Analysis (EQ3)}

\noindent\ul{\textbf{Analysis 1: Additional Results of Source Family Prediction.}}
Beside enhancing the learning of LLMGT detection head, the introduction of a family classifier makes it possible to provide a source family prediction simultaneously, which facilitates further forensics analysis and may improve users' trustworthiness to the detector. Table~\ref{tab:family-pred-results} previews the current performance of known family prediction. The performance is mixed. The highest F1 score exceeds 97\% while the lowest is below 50\%, 
demonstrating that the family prediction is far more challenging then LLMGT detection and the current design requires improvement for family prediction. Recalling the main results, this also implies that even a moderate family-aware learning could bring performance improvement, indicating that its potential remains to be further explored.

\begin{table}[ht]
    \small
    \centering
    \caption{\label{tab:family-pred-results} F1 scores for family classification of unseen fine-tuned LLM-generated texts.}
    \begin{tabular}{@{}clcccc@{}}
        \toprule
        \multicolumn{2}{c}{\textbf{Dataset}} & \textbf{LLaMA} & \textbf{Mistral} & \textbf{Gemma} & \textbf{Macro} \\
        \midrule
        \multirow{2}{*}[0pt]{arXiv} & Full & 73.59 & 72.44 & 97.81 & 81.28 \\
        & LoRA & 73.17 & 85.43 & 59.11 & 72.57 \\
        \midrule
        \multirow{2}{*}[0pt]{Q\&A} & Full & 73.68 & 55.56 & 73.45 & 67.56 \\
        & LoRA & 47.56 & 53.91 & 81.82 & 61.10 \\
        \bottomrule
    \end{tabular}
\end{table}

\noindent\ul{\textbf{Analysis 2: Comparison with Commercial Detectors.}}
We compare PhantomHunter with four commercial detectors that supports API calling on the whole test set. We assume that commercial detectors have learned on text generated by the base models of involved families and been trained on a larger and more diverse training corpus than ours. However, Table~\ref{tab:commercial-results} shows that PhantomHunter, trained on a small but focused dataset instead, significantly outperforms the compared ones, again highlighting the value of the proposed family-aware learning.

\begin{table}[ht]
\small
\centering
\caption{\label{tab:commercial-results} Performance comparison with commercial detectors (\%). *We remove samples with <300 characters for WinstonAI due to its minimum length constraint.}
\begin{tabular}{lcccc}
\toprule
\textbf{Method/Service} & \textbf{TPR\textsubscript{1\%FPR}} & \textbf{F1\textsubscript{Human}} & \textbf{F1\textsubscript{Gen}} & \textbf{F1\textsubscript{Macro}} \\
\midrule
WinstonAI* & 32.81 & 27.48 & 87.29 & 57.39 \\
Sapling & 44.85 & 43.56 & 93.27 & 68.42 \\
BlueEyes & 59.26 & 35.85 & 89.24 & 62.55 \\
HasteWire & 69.90 & 65.01 & 97.50 & 81.26 \\
\midrule
\textbf{PhantomHunter} & \textbf{99.45} & \textbf{93.43} & \textbf{99.56} & \textbf{96.50} \\
\bottomrule
\end{tabular}
\end{table}

\noindent\ul{\textbf{Analysis 3: A Practice of Including Seen LLMs.}}
In practice, LLMGT detectors should maintain good performance against both seen and unseen LLMs. Here we show that PhantomHunter can be easily extend to be compatible to other seen LLMs through a simple modification.
Specifically, an additional category ``others'' is added to the label space of the family classifier $\mathrm{MLP_F}$, which is used as the family label of these texts from other seen models, and the number of experts in MoE is increased accordingly.
Table~\ref{tab:others-seen-results} shows that PhantomHunter maintains high performance for not only unseen LLMs but also the seen GPT-4o mini\footnote{\url{https://openai.com/index/gpt-4o-mini-advancing-cost-efficient-intelligence/}} and Claude 3.7 Sonnet\footnote{\url{https://www.anthropic.com/news/claude-3-7-sonnet}}, which indicates its good compatibility and extendibility to more diverse LLMs.

\begin{table}[ht]
\small
\centering
\caption{\label{tab:others-seen-results} Performance comparison of seen and unseen fine-tuned LLM-generated texts (\%), after adding the "others" family category.}
\begin{tabular}{clcccc}
\toprule
\multicolumn{2}{c}{\textbf{Model}} & \textbf{TPR\textsubscript{1\%FPR}} & \textbf{F1\textsubscript{Human}} & \textbf{F1\textsubscript{Gen}} & \textbf{F1\textsubscript{Macro}} \\
\midrule
\multirow{3}{*}[0pt]{\rotatebox{90}{Unseen}} & LLaMA-cs & 99.42 & 91.67 & 99.24 & 95.45 \\
& Mistral-cs & 96.80 & 91.67 & 99.21 & 95.44 \\
& Gemma-cs & 91.67 & 90.72 & 98.84 & 94.78 \\
\midrule
\multirow{2}{*}[0pt]{\rotatebox{90}{Seen}} & GPT-4o mini & 92.67 & 91.67 & 97.40 & 94.53 \\
& Claude 3.7 Sonnet & 80.67 & 91.19 & 97.91 & 94.55 \\
\bottomrule
\end{tabular}
\end{table}

\section{Related Works}

\noindent\ul{\textbf{LLM-Generated Text Detection.}}
Existing methods can be categorized as model-based and metric-based ones. The former generally trains neural models to analyze the textual characteristics of LLM outputs, exploiting semantic features~\cite{Guo2023,guo2024detective,yu2024text} and self-familiarity~\cite{ji2025iknowbetterreally}. The latter often sets similarity metrics based on LLMs' output probability or behavioral characteristics~\cite{zhu-etal-2023-beat,mitchell2023detectgpt,hans2024spotting,yu2024dpic}. 
However, existing methods and evaluation benchmarks~\cite{he2024mgtbench,wu2024detectrl,wang2024m4gt} focuses on publicly known LLMs, leading to the ignorance of detectors' unsatisfying performance that we experimentally showed.

One possible solution is to add watermarks to base models' outputs~\cite{liu2024survey}, but it requires LLM providers to pay extra costs and recent research warns that LLM watermarks could be distorted or even removed through various adversarial techniques~\cite{wu2024bypassing,pang2024no}.
PhantomHunter showcases a non-watermarking solution to improve detection for text from unseen privately-tuned LLMs.

\noindent\ul{\textbf{Family Analysis of LLMs.}}
Analyzing commonalities and uniqueness of LLMs from a family perspective yields a deep understanding of LLMs' relationship~\cite{kumarage2024survey}, which is useful for LLM attributions~\cite{10438784,poger} and LLMs' license violation detection~\cite{foley-etal-2023-matching}. As shown by~\citet{sarvazyan2023supervised,thorat2024}, the family factor could influence text detectability and detectors' generalizability.
However, they only investigated the base models of different sizes, ignoring the unseen privately-tuned LLMs that we focuses on in this research.

\section{Conclusion and Future Work}
We proposed PhantomHunter, which is specialized to improve detection performance of text generated by unseen privately-tuned LLMs. We first experimentally revealed that fine-tuning open-source base models would significantly lower detectability of generated text, and then addressed this issue via family-aware learning to capture shared traits in each LLM family. Comparison experiments with seven methods and three industrial services confirmed PhantomHunter's superiority. The further analysis exhibited that PhantomHunter has good compatibility and extendibility to better adapt to real-world situations. We plan to enhance its generalizability to unknown LLM families and lower its cost of memory in the future.

\section*{Ethical Considerations}

{\textbf{Ethical Concerns Regarding Technical limitations.}} Though PhantomHunter showcases how to tackle the performance degradation issue when encountering texts generated by privately-tuned LLMs based on widely-known open-source ones, we also identify the following limitations of PhantomHunter which may influence the real-world implementation. First, the current version only supports binary classification between human-written and LLM-generated text plus the possible family source for a suspicious LLM-generated text piece. It does not provide fine-grained sentence- or token-level annotations~\cite{wang2023seqxgpt} and the current performance of identifying family sources is moderate. Second, due to the experimental cost constraint, our experiment only covered the several common LLM families and the performance on other families like Qwen~\cite{Qwen} and DeepSeek~\cite{deepseekr1,deepseekv3} remains unknown. Third, extracting features from base LLMs requires them to be locally deployed and thus increases the memory cost and the inference computation overhead, which trades computation costs for higher detection performance.

Therefore, these directions requires further exploration and we advocate more industrial solutions to improve the trustworthiness and efficiency of LLM-generated text detection systems. Meanwhile, it is not recommended that the feedback of PhantomHunter serves as a sole basis for real-world actions (e.g, disciplinary action against any student). Additional harmful content detection~\cite{hu2024bad,sharma2025constitutional} and human verification~\cite{niu-etal-2024-detecting} are needed.

\noindent{\textbf{Data Collection.}} We did not collect any private personal data in our experiment. The data we used is from an existing dataset~\cite{Guo2023} and a public preprint server. We do not find sensitive, non-anonymous personal data in it.

\noindent{\textbf{Disclaimer of the Evaluation on Commercial Detectors.}} We test several commercial detectors on an extremely challenging scenario where the open-source LLMs are privately fine-tuned and deployed to generated text. The authors are not intended to conduct a comprehensive and unbiased evaluation that could reflect any key advantages of these services and we do not imply any further judgments beyond the analysis in this paper.

\appendix

\section{Supplementary Experimental Details}
\label{app:1}

\subsection{Further Introduction of Fine-tuning Data}
\begin{itemize}
    \item \textbf{arXiv}: We collected titles and abstracts of academic papers in various domains from its archive, including computer sciences, physics, quantitative biology, statistics, electrical engineering and systems science, mathematics, quantitative finance, economics. It covers the eight main subjects on arXiv and results in a set of 69,297 samples.
    \item \textbf{Q\&A}: We obtained question-answering data from the HC3 dataset, which provides human-written data across various domains aggregated from Q\&A benchmarks in Chinese and English. In this paper, we select data of three domains, including ELI5 (long answers from Subreddit \textit{Explain Like I’m Five};~\citealp{eli5}), finance (from FiQA dataset;~\citealp{FinQA}), and medicine (from MedDialog dataset;~\citealp{zeng2020meddialog}).
\end{itemize}

\subsection{Prompts and Cases in Data Generation}

We adopt the two following prompt templates for data generation: 

\begin{tcolorbox}[title=Prompt 1: Generating an abstract given the title of an academic paper, boxrule=0pt, left=1mm, right=1mm, top=1mm, bottom=1mm, fontupper=\small, colback=cyan!10]
Write an abstract for the academic paper titled [\emph{title}].
\end{tcolorbox}

\begin{tcolorbox}[title=Prompt 2: Generating an answer given a question, boxrule=0pt, left=1mm, right=1mm, top=1mm, bottom=1mm, fontupper=\small, colback=cyan!10]
[\emph{question}]
\end{tcolorbox}

\subsection{Details of Fine-tuning Methods}

In this work, we utilize the LoRA (Low-Rank Adaptation) and Full fine-tuning techniques for fine-tuning the pretrained model. The key hyperparameters for the fine-tuning process are detailed as follows:

\begin{itemize}
    \item \textbf{Full fine-tuning} was performed at the \texttt{sft} (supervised finetuning) stage. The training configuration was set with a batch size of 1 per device and gradient accumulation steps of 4. The learning rate was set to 1.0e-5, and the maximum number of training epochs was 10. A cosine learning rate scheduler was used with a warm-up ratio of 0.1. BF16 precision was enabled for efficient training, and the DDP (Distributed Data Parallel) strategy was employed.
    \item \textbf{LoRA fine-tuning}~\cite{lora} was performed with the LoRA rank set to 8, and the training was conducted with a batch size of 1 per device, gradient accumulation steps of 4, a learning rate of 1.0e-5, and maximum number of epochs of 10, utilizing a cosine learning rate scheduler with a warm-up ratio of 0.1, BF16 precision for efficient training.
\end{itemize}

\subsection{Introduction of Compared Commercial Detectors}
We select four commercial detectors for comparison with industrial products. Due to the inaccessibility, we do not consider the commercial detectors with the requirement of additional actions besides API callings (e.g., Grammarly requires a subscription of the enterprise or education plan first\footnote{\url{https://developer.grammarly.com/ai-detection-api.html}}) or have strict constraints in request times (e.g., Tencent's Zhuque AI Detector allows only 20 requests per day\footnote{\url{https://matrix.tencent.com/ai-detect/}}). We check the accessibility of APIs in 2025. The introduction of the selected detectors are as follows:

\begin{itemize}
    \item \textbf{WinstonAI (\url{https://gowinston.ai/}):} Due to its minimum length constraint of 300 characters, we remove the text samples shorter than 300 characters for WinstonAI's API requests, resulting in a smaller testing set of 6,492 samples (97.77\% of the whole). Considering shorter text is generally harder to distinguish~\cite{tianmultiscale}, the removal may bring WinstonAI detector an extra advantage over others.
    \item \textbf{Sapling (\url{https://sapling.ai/}):} Sapling AI Content Detector uses a machine learning system (a Transformer network) similar to that used to generate AI content. The officially reported detection rate for AI-generated content is over 97\% with less than 3\% false positive rate for human-written content. 
    \item \textbf{HasteWire (\url{https://hastewire.com/humanizer/detect}):} An AI solutions provider for businesses that specializes in building custom AI solutions. It provides AI content detector which leverages advanced machine learning models to provide a confidence score on content origins. Its model has been trained against models such as GPT-4o, Claude 3.5 Sonnet, DeepSeek-Chat, Llama-3.2, Mistral Small 2501, o3-mini and more.
\end{itemize}

\bibliography{main}

\end{document}